\documentclass[conference]{IEEEtran}

\usepackage{graphicx}
\usepackage{colortbl}
\usepackage{amssymb}
\usepackage{amsmath}
\usepackage{amsthm}
\usepackage{balance}
\usepackage[T1]{fontenc}
\usepackage[ruled,lined,linesnumbered,commentsnumbered]{algorithm2e}
\usepackage{url}
\usepackage{tabularx}

\SetKwInput{Input}{Input}
\SetKwInput{Output}{Output}
\SetAlgoSkip{smallskip}

\begin{document}

\title{Helping AI to Play Hearthstone:\\ AAIA'17 Data Mining Challenge}

\author{
\IEEEauthorblockN{Andrzej Janusz\IEEEauthorrefmark{1}\IEEEauthorrefmark{2}, Tomasz Tajmajer\IEEEauthorrefmark{1}\IEEEauthorrefmark{2}}
\IEEEauthorblockA{
    \IEEEauthorrefmark{1}Institute of Informatics, University of Warsaw\\
    Banacha 2, 02-097 Warsaw, Poland\\
    {\{janusza,t.tajmajer\}@mimuw.edu.pl}} \and
\IEEEauthorblockN{Maciej \'Swiechowski\IEEEauthorrefmark{2}}
\IEEEauthorblockA{
    \IEEEauthorrefmark{2}Silver Bullet Solutions\\
    Liwiecka 25, 04-289 Warsaw, Poland\\
    {m.swiechowski@mini.pw.edu.pl}} \and
}

\maketitle

\begin{abstract}
This paper summarizes the AAIA'17 Data Mining Challenge: Helping AI to Play Hearthstone which was held between March 23, and May 15, 2017 at the Knowledge Pit platform. We briefly describe the scope and background of this competition in the context of a more general project related to the development of an AI engine for video games, called Grail. We also discuss the outcomes of this challenge and demonstrate how predictive models for the assessment of player's winning chances can be utilized in a construction of an intelligent agent for playing Hearthstone. Finally, we show a few selected machine learning approaches for modeling state and action values in Hearthstone. We provide evaluation for a few promising solutions that may be used to create more advanced types of agents, especially in conjunction with Monte Carlo Tree Search algorithms.

\end{abstract}

\begin{IEEEkeywords}
data mining competition; AI in video games; MCTS; artificial neural networks; Hearthstone: Heroes of Warcraft;
\end{IEEEkeywords}

\IEEEpeerreviewmaketitle

\section{Introduction}\label{sec_intro}

\IEEEoverridecommandlockouts\IEEEPARstart{H}{earthstone: Heroes of Warcraft} is a free-to-play online video game developed and published by Blizzard Entertainment. 
It is an example of a turn-based collectible card game played between two opponents. During a game, players use their custom decks of thirty cards, along with a selected hero with a unique power. They spend mana points to cast spells or summon minions to attack the opponent, with the goal to reduce the opponent's health to zero. Building efficient decks is an essential skill and many archetypes of decks exists. These archetypes are characterized by different distributions of the cards' mana cost and thus are meant for players with different play styles. There are also sets of cards, which synergize well due to their unique properties and can be used in many different decks.

In recent years, Hearthstone has become a testbed for AI research. A community of passionate players and developers have started the HearthSim project (\url{https://hearthsim.info/}) and created many tools that allow simulating the game for the purpose of AI and machine learning experiments. Several researchers have already used this game in their studies~\cite{taralla2015learning,garcia2016evolutionary}. Moreover, our research team decided to use Hearthstone as one of case studies which aim to demonstrate capabilities of our video game's AI designing framework, called Grail. For this reason, one objective of this article is to explain how some powerful heuristic search algorithms can be combined with prediction models that derive from the machine learning domain, in order to construct a smart and cunning artificial Hearthstone player.

The paper is organized as follows: in the next section, we describe the specificity of the AAIA'17 Data Mining Competition. In Section~\ref{sec_expAnalysis}, the approach of using the collected data to effectively play the game of Hearthstone is presented. The approach is based on the so-called Monte Carlo Tree Search algorithm (Section~\ref{sec_mcts}) coupled with machine learning models (Sections~\ref{sec_mcts_heuristics}~--~\ref{sec_lstm}). The last section is devoted to conclusions.

\section{AAIA'17 Data Mining Challenge}\label{sec_aaia16dmc}

AAIA'17 Data Mining Challenge: Helping AI to Play Hearthstone (\url{https://knowledgepit.fedcsis.org/contest/view.php?id=120}) took place between March 23 and May 15, 2017. It was organized under the auspices of the $12^{th}$ International Symposium on Advances in Artificial Intelligence and Applications (AAIA'17, \url{https://fedcsis.org/2017/aaia}) which is a part of the FedCSIS conference series.

The main objective in this competition was to construct a prediction model which would be able to foresee who is going to win, using only information about a single game state. The ability to accurately assess winning chances of a player in different game states is substantial for designing efficient and challenging AI opponents in many video games. The most famous example is the AlphaGo program, which used two neural networks to evaluate possible moves and game states of Go games~\cite{alphaGo}. In our competition, we challenged participants with the task to design such models for Hearthstone.  

In particular, the dataset provided to participants contained examples of game states extracted from Hearthstone play outs between weak AI players (i.e. the agents which were used to generate the data were choosing their in-game decisions at random). The participants were asked to predict winning chances of the first player from game states belonging to the test set and submit their predictions to the Knowledge Pit competition platform \cite{DBLP:conf/csp/JanuszSSR15}. In order to give participants a freedom of choosing a representation of the data which they want to use, the datasets were provided in two formats: in a tabular format (with simplified representation) and as raw JSON files (with detailed game states).

The training part of the data was made available along with the corresponding information regarding the actual game winners. These labels were removed from the test set which was also made available to participants. Initially, the training set consisted of $2000000$ game states, however, after detecting an unwitting data leakage \cite{KaufmanRPS12}, after first few weeks of the challenge, it was extended by additional $1250000$ cases from the original test set (in total, there were $3250000$ training examples). The final test set consisted of $750000$ game states. Test set examples were obtained from a different set of Hearthstone play outs than the training cases. In fact, while the training data contained game states from $\approx 65000$ simulations, more than $180000$ play outs were simulated to generate the test set. It is also important to note that while in the training games there were used only $9$ different sets of cards (one deck for every hero type), the test games were played using $27$ different decks. As a consequence, the test data contained Hearthstone cards which had never appeared in the training set. Table \ref{tab_dataOverview} shows a summary of basic characteristics of datasets used in the challenge.

\begin{table}[t]
	\caption{Basic characteristics of datasets used in AAIA'17 Data Mining Challenge.}\label{tab_dataOverview}
	\begin{center}
		\begin{tabular}{p{2.5cm}|p{1.9cm}|p{1.9cm}}
			\hline 
			characteristic & training set & test set\\
			\hline
			no. examples & $3250000$ & $750000$\\
			no. games & $65000$ & $180000$\\
			no. used decks & $9$ & $27$\\
			percent of wins & $50.46\%$ & $57.27\%$\\
			min. win rate per hero (percent/hero\_id) & $50.07\%/326$ & $37.53\%/754$\\
			max. win rate per hero (percent/hero\_id) & $50.85\%/981$ & $75.34\%/25$\\
			\hline
		\end{tabular}
	\end{center}
\end{table}

\subsection{Evaluation of results and participation in the challenge}\label{sec_evaluation}

Participants of the competition had to prepare their solutions in a form of a file with predictions of a likelihood that \textit{player 1} will win, given a corresponding description of a game state. The files with predictions had to be sent using the submission system of Knowledge Pit \cite{DBLP:conf/csp/JanuszSSR15}. Each of the competing teams could submit multiple solutions. Quality of the submissions was measured using Area Under the ROC Curve (AUC) \cite{hastie01statisticallearning}. The submitted solutions were evaluated on-line and the preliminary results were published on the competition leaderboard. The preliminary score was computed on a subset of the test set, fixed for all participants. Size of this subset corresponded to randomly chosen 5\% of the test set. The final evaluation was conducted after completion of the competition using the remaining part of the test data.

Apart from submitting their predictions, each team was also obligated by competition rules to provide a brief report describing its approach. Only the final solutions from teams which sent a valid report could undergo the final evaluation and be published among the competition results. In this way, we were able to collect a vast amount of information regarding efficient representation methods of Hearthstone game states and state-of-the-art approaches to this type of prediction problems.


\subsection{Summary of the competition}\label{sec_resultsOverview}

Even though AAIA'17 Data Mining Challenge lasted for less than two months, it attracted attention of many researchers from domains of machine learning and artificial intelligence in video games. By the end of competition there were $296$ teams from $28$ countries registered in the challenge. Among them, $188$ teams submitted at least one solution to the leaderboard and $114$ teams described their solution in a report uploaded to the Knowledge Pit platform. In total, we received $4067$ submission, which makes this competition the most popular one among challenges organized at Knowledge Pit to~this~day.

The large number of submitted reports gave us a unique opportunity to review the most effective prediction methods for the assessment of Hearthstone game states. The most successful approach in this regard turned out to be artificial neural networks \cite{DBLP:journals/tcs/SzczukaS11} and particularly, the deep learning methods \cite{Krizhevsky:2012:ICD:2999134.2999257}. In fact, all top-ranked teams used neural networks in their solutions and the winners focused particularly on the convolutional neural networks \cite{Karpathy:2014:LVC:2679600.2680211}.  Another popular approach was the utilization of \textit{xgboost} algorithm \cite{Chen:2016:XST:2939672.2939785}. There were also much simpler approaches which turned to be efficient, such as the logistic regression models. Moreover, all of these methods were often combined -- techniques such as averaging, bagging or stacking were commonly used to obtain better prediction results \cite{Janusz12ida}. Table \ref{tab_compResults} presents scores obtained by the five top-ranked teams. Noticeable is the fact that the difference in scores between the best solution and the baseline is~less~than~2\%.

\begin{table}[t]
	\centering
	\caption{Final results and number of submissions from the top ranked teams. The last row shows the result obtained by our baseline solution -- a fully-connected neural network with two hidden layers, trained on the tabular data.}\label{tab_compResults}
	\begin{tabular}{p{2.0cm}|l|p{1.1cm}|l}
		\hline
		team name & rank & \mbox{\# of} \mbox{submissions} & final result \\ 
		\hline
		iwannabetheverybest &     $1$ & $139$ & $0.8019$ \\ 
		hieuvq &     $2$ & $384$ & $0.7992$ \\ 
		johnpateha &     $3$ & $143$ & $0.7990$ \\ 
		vz &     $4$ & $11$ & $0.7973$ \\ 
		jj &     $5$ & $75$ & $0.7971$ \\ 
		$\cdots$ & $\cdots$ & $\cdots$ & $\cdots$ \\
		baseline & $94$ & -- & $0.7846$ \\
		\hline
	\end{tabular}
\end{table}

Many teams decided to use data in the JSON format in order to construct richer representations of game states than the one which was available in the provided tabular data. Feature engineering \cite{DBLP:journals/isf/GruzdzIS06} turned out to be an important aspect of the most efficient solutions. Extracted features were often a reflection of participant's experience and domain knowledge about Hearthstone. Their descriptions included in reports turned out to be a valuable source of knowledge which can be used to improve our artificial Hearthstone players.

\section{Augmentation of game state search heuristics with neural networks}\label{sec_expAnalysis}

\subsection{Monte-Carlo Tree Search}\label{sec_mcts}

\begin{figure}[t]
\begin{center}
\includegraphics[scale=0.5]{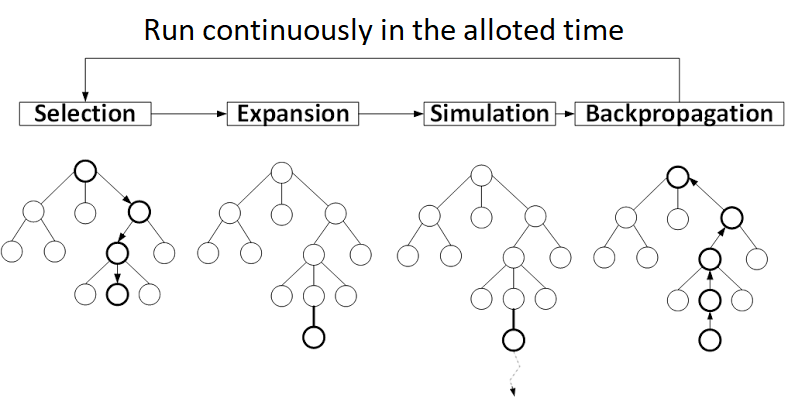}
\vspace{-0.2cm}
\caption{Depiction of four phases which comprise the MCTS algorithm.}
\label{fig:mcts}
\end{center}
\end{figure}

Monte Carlo Tree Search (MCTS)~\cite{mctsSurvey} is a method of learning an optimal policy for solving problems such as game-playing. For the first time, it was used for games in Go~\cite{gelly2006} as an improvement over a Monte Carlo sampling technique (without the tree search). The algorithm led to a breakthrough in the game of Go, which had been previously regarded as intractable for computer programs~\cite{goChallenge}. Driven by this success, MCTS became the state-of-the-art approach in various game domains, such as General Game Playing~\cite{swiechowski2015recent} and General Video Game Playing~\cite{perez2014knowledge}. The idea of MCTS is to repeatedly simulate the game (problem) and build statistics about states and actions. Each iteration of the algorithm consists of four phases as depicted in Figure~\ref{fig:mcts}.

\textbf{(1) Selection.} In this phase, the algorithm starts from the root node and searches the tree down by choosing subsequent children nodes. The child node at each node down the path is chosen according to the so-called selection policy. The selection phase ends when there is no child node to choose, i.e., a leaf node has been reached.\\
\indent \textbf{(2) Expansion.} One of the possible actions is applied to a node selected in the previous step and the tree is grown by adding a child node representing the resulting state.\\
\indent \textbf{(3) Simulation.} The algorithm starts from the new node and performs a complete game simulation, i.e., reaching a terminal state. This phase is done outside the game-tree and no nodes are added to it. Once the simulation reaches the terminal state, the obtained goals (outcomes) of each player are fetched.\\
\indent \textbf{(4) Back-propagation.} Here, the statistics are recalculated inside all nodes along the path from the root to the leaf (containing the starting state for the simulation) in the game tree. The statistics include the average scores of each player and the number of visits to a node. An average score is computed as the total score achieved in iterations going through a particular node divided by the number of visits to that node.

In the classic implementation, actions in the simulation phase are chosen with respect to uniform random distribution. In the selection phase, a more sophisticated formula (selection policy) is typically used. The most common one, which was also employed in this paper for all MCTS-based programs used during the experiments is called Upper Confidence Bounds applied for Trees (UCT)~\cite{uct}.
\begin{equation}
a^{*} = \arg\max_{a \in A(s)}\left \{ Q(s,a)+C\sqrt{\frac{ln\left
[N(s)  \right ]}{N(s,a)}} \right \}\label{eq:uct}
\end{equation}
where $A(s)$ is a set of actions available in state $s$, $Q(s,a)$ denotes the average result of playing action $a$ in state $s$ in the simulations performed so far, $N(s)$ - a number of times state $s$ has been visited in previous simulations and $N(s,a)$ - a number of times action $a$ has been sampled in this state in previous simulations. Constant $C$ controls the balance between exploration and exploitation. 

The MCTS algorithm using the UCT selection formula is proved to converge to the min-max theoretical optimum~\cite{uct}. However, it poses several advantages over a classic min-max search. For instance, it does not require any game specific evaluation function and constructs the tree in an asymmetric manner, focusing at the most promising lines of play. It scales better with the depth of the tree, it can be stopped at anytime to return the best action found so far. 

\subsection{Monte-Carlo Tree Search with Heuristics}\label{sec_mcts_heuristics}

Despite the wide usage in a variety of game domains, the MCTS method has bottlenecks and limitations. It is both computationally demanding and memory intensive. Games with huge branching factor, i.e., the total number of actions available to players, in average, often inhibit the usage of MCTS and other tree-search methods. This weakness has motivated us to combine this algorithm with heuristics represented by prediction models. Such prediction models can be trained to either predict the outcome of the game by looking at a potential next state (candidate state) of the game or at a potential action (candidate action). In the scope of this paper, we will use the terms ``machine learning prediction models'' and ``heuristic evaluation'' interchangeably.   

There is a couple of ways to combine external heuristics with the MCTS algorithm. The authors of paper~\cite{magician} give a nice review of four common methods:

\textbf{(1) Tree Policy Bias} - here the heuristic evaluation function is included together with the $Q(s,a)$ in the UCT formula (see Eq.~\ref{eq:uct}) or its equivalent. A typical implementation of this idea is called \textit{Progressive Bias}~\cite{chaslot2008progressive}, in which the standard UCT evaluation is linearly combined with the heuristic evaluation with the weight proportional to the number of simulations. The more simulations are performed, the more statistical confidence, and therefore, the higher weight is assigned to the standard UCT formula.\\
\indent \textbf{(2) Move Ordering} - the heuristic defines the order, in which actions in the tree are expanded (chosen for the first time). This method has the most significant impact on the deeper parts of the tree, because the MCTS is less likely to visit them again, so the order matters. If better moves are expanded first, their neighbourhood in the tree has a higher chance to be visited in subsequent simulations.\\
 \indent \textbf{(3) Simulation Policy Bias} - in the baseline version of the MCTS algorithm, the actions during the simulation phase are chosen randomly. With a good heuristic evaluation, a sensible approach is to infer this knowledge in the action selection process, while still leaving some degree of randomness. The two most common implementations are pseudo-roulette selection with probabilities computed using Boltzmann distribution (where the heuristic evaluation is used) or the so-called epsilon-greedy approach~\cite{miniplayer-tciaig}. In the latter, the action with the highest heuristic evaluation is chosen with the probability of $\epsilon$ or a random one with the probability of $1-\epsilon$.\\
\indent \textbf{(4) Early Cutoff} - the authors of~\cite{magician} achieved the best results with terminating Monte Carlo simulations before the game ends and returning the heuristic evaluation of the current state. This variant is called Early Cutoff and the cutoff is done typically at fixed depth or with certain small probability (e.g. P=$0.1$) in each step. 

The aforementioned AlphaGo program employs both, Tree Policy Bias and Simulation Policy Bias. Motivated by its success, we decided to apply a similar approach for Hearthstone.

\subsection{Generality of models trained on random simulations}\label{sec_generality}

Both Tree Policy Bias and Simulation Policy Bias methods utilize a heuristic function which provide the value of a game state or an action. Various machine learning methods may be used to obtain these evaluations, including supervised prediction models \cite{hastie01statisticallearning}. Since random simulations are used in the classic version of MCTS, it is natural to train such models on game states obtained from play outs between agents making random decisions.

\begin{figure}[t]
\begin{center}
\includegraphics[width=0.42\textwidth,height=0.45\textwidth]{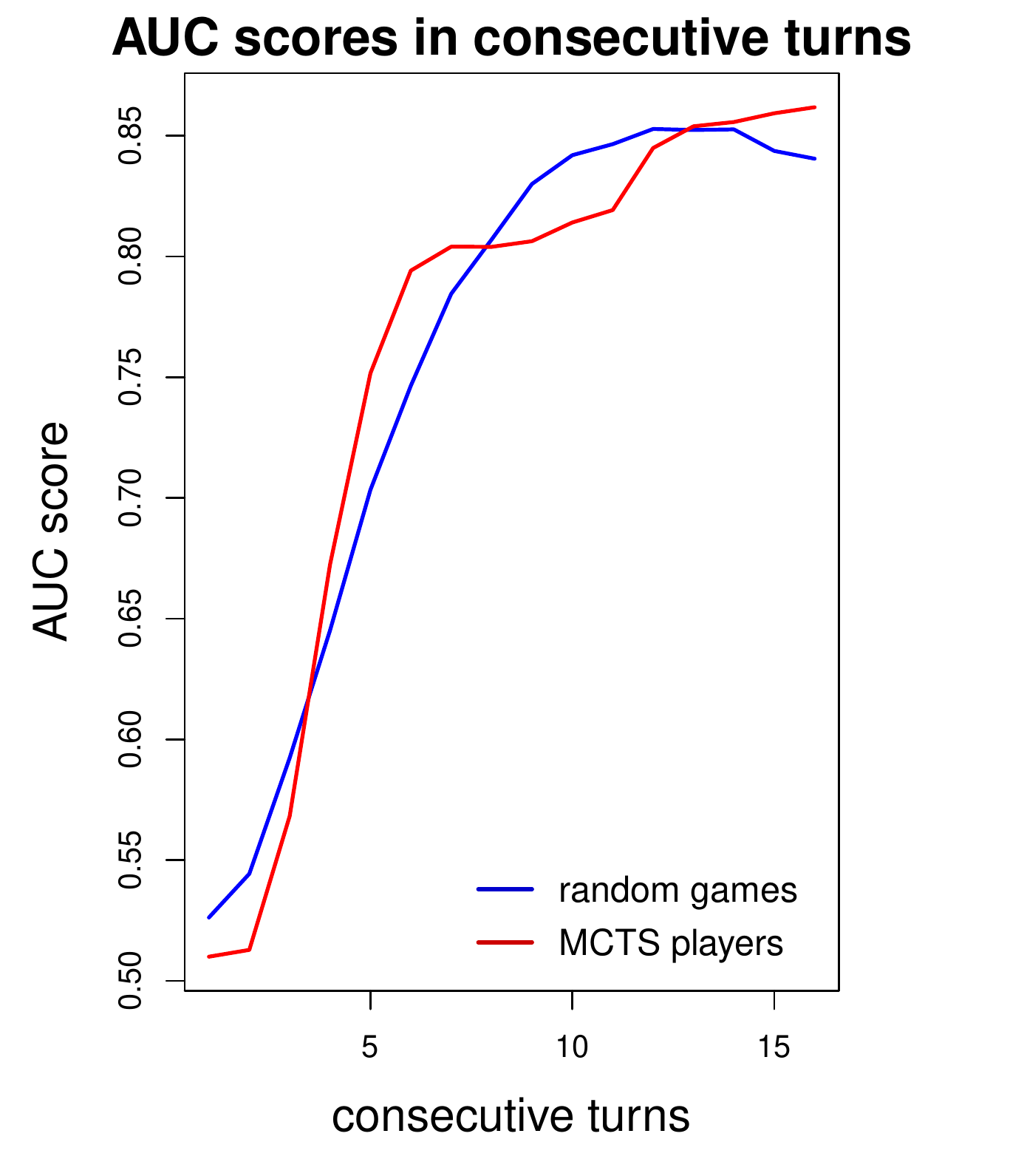}
\vspace{-0.2cm}
\caption{A comparison between performance of a neural network trained on AAIA17 Data Mining Challenge dataset when tested on game states generated by random and MCTS agents.}
\label{fig:random_vs_mcts}
\end{center}
\end{figure}

The datasets used in AAIA17 Data Mining Challenge constitute an example to this type of data. It can be used to train prediction models for the purpose of evaluation of Hearthstone game states during MCTS simulations. However, a question arises whether such models could be also effective in games played by more intelligent opponents. To check that, we conducted a series of experiments. First, we generated an additional dataset containing game states from duels between strong MCTS agents. Each agent was making decisions after performing $15000$ random simulations before a single action. In total, there were $28604$ play outs generated in this way, which resulted in a dataset consisting of $814407$ game states. We trained a simple neural network with two hidden layers on the training set used in our competition and we checked its performance on the available test set. Next, we constructed another model using all competition data and we tested it on the additional dataset. Figure \ref{fig:random_vs_mcts} shows a comparison of the obtained AUC scores in consecutive turns. Surprisingly, total AUC dropped only slightly (from $\approx 0.79$ to $\approx 0.75$) when the test was done on the data generated by MCTS players. It shows that predictive models can be successfully used for evaluation of game states, even in a case when they are trained on random simulations.

\subsection{Learning a playing strategy from sequences}\label{sec_lstm}

In practice, it is often desirable to have a function that provides the policy rather than the value of a particular action. The policy $\pi(a|s)$ specifies the probabilities for all actions available in a given state, thus it enables the selection of the best action candidate in a single state evaluation.

Reinforcement Learning is used in particular for cases where the optimal policy is unknown and needs to be learned based on sparse reward signals. However, a supervised learning approach may be used, when examples of policies are available (e.g. from human players or other algorithms). In our case, we may use MCTS to generate Hearthstone matches and obtain state-action pairs i.e. record what action was chosen by MCTS as a response to given game state. Next, we may train a model that predicts the action that MCTS would choose for a given state. Training such a model is basically a classification task, well fitted for deep neural networks (DNN).

Long short-term memory (LSTM) is a type of recurrent neural network \cite{LSTM} dedicated for use with sequences. The architecture of LSTM enables it to learn long and short term temporal dependencies. LSTM provides superior performance in tasks such as speech recognition, machine translation or language modeling. A deep LSTM network may be created by stacking multiple LSTM layers.

A DNN may be trained to approximate a policy from examples of state-action pairs - we will refer to this network as to \textit{policy network} \cite{alphaGo}. However, in case of Hearthstone, a single state may not provide enough information to the model for a valid action prediction. This is due to the fact that a single turn in Hearthstone consists of many moves. Moreover, a single move is decomposed into a few atomic actions in order to be efficiently implemented in the Hearthstone simulator. For example, putting a minion on the board consists of two actions: selecting a card from hand and selecting the slot on the board where the minion should be placed. To improve the accuracy of the policy predictions, rather than using a single state, we chosen to use a sequence of states and previous actions as the input to the policy network.

\begin{figure}[t]
\begin{center}
\includegraphics[scale=1.5]{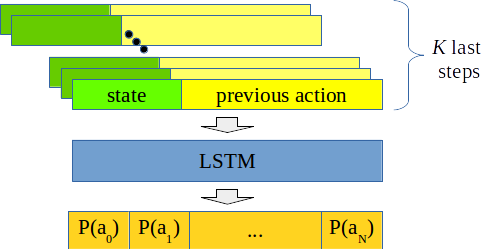}
\caption{LSTM policy network. A sequence of states and actions from previous game steps are provided as the input to the LSTM network. The LSTM network is trained to output probabilities for actions available after the last step in the sequence.}
\label{fig:lstm}
\end{center}
\end{figure}

Our policy network is presented in figure \ref{fig:lstm}. LSTM network is provided with a sequence of $K=10$ vectors created from concatenating a state vector for state $s_{t-k}$ and an action vector for action $a_{t-k-1}$, for $k \in [0, 1, ...,  K-1]$. The state vector includes 403 values representing the state of the game from the point of view of a selected player. The action vector is of length 91 and contains a one-hot encoded action. The output of the LSTM is a single vector with probabilities assigned to each of all available 91 actions (the probabilities are provided also for actions that are illegal in a certain state). The LSTM in our experiments consisted of 3 layers of 256 LSTM cells with dropout

To obtain the training data, we first used MCTS with 1000 iterations to generate 7000 games between randomly selected decks. Using this data we obtained 528365 sequences of length 10 where each element of the sequence included 494 values. We will refer to this dataset as to \textit{SeqMCTS1k}. Next, we generated 1500 games using MCTS with 10000 iterations. As a result we created a second dataset: \textit{SeqMCTS10k}, that consists of 149521 sequences.

To evaluate the policy network, we created a greedy DNN agent that always selects the most promising action from the predictions of the policy network. We confronted this agent against a random agent and an agent using MCTS with 1000 iterations. Greedy DNN agents were using policy networks trained in three variants: 1) using only \textit{SeqMCTS1k} dataset, 2) using only \textit{seqMCTS10k} dataset and 3) trained first on the \textit{SeqMCTS1k} dataset and then retrained on the \textit{SeqMCTS10k} dataset. The results are presented in Table \ref{tab_lstm_eval}. Each score is calculated based on 500 games played between~the~agents.

\begin{table}[t]
	\caption{Evaluation results of agents using LSTM policy network. }\label{tab_lstm_eval}
	\begin{center}
		\begin{tabular}{p{2.2cm}|p{2.2cm}|p{2.2cm}}
			\hline
			greedy agent type & wins vs random agent & wins vs MCTS(1000)  \\
			\hline
			seqMCTS1k & $96.2\%$ & $23.0\%$ \\
            seqMCTS10k & $98.6\%$ & $26.4\%$ \\
            seqMCTS1k retrained with seqMCTS10k & $98.2\%$ & $50.4\%$ \\
			\hline
		\end{tabular}
	\end{center}
\end{table}

\section{Conclusions}\label{sec_conclusions}

In the paper we provided a summary of AAIA'17 Data Mining Challenge which was held at the Knowledge Pit platform. Results of this competition clearly show that learning from Hearthstone game logs is feasible and has a potential to facilitate a construction of intelligent artificial agents which play that game. We explained how prediction models can be combined with game state search heuristics to improve their performance. We also demonstrated results of experiments showing that models trained on data obtained from random simulations can be successfully applied for the assessment in games between intelligent agents. Finally, we showed that more advanced approaches such as the supervised action policy learning based on game state sequences are feasible and deserve further investigation.

\section*{Acknowledgments}

This research was co-funded by the Smart Growth Operational Programme 2014-2020, financed by the European Regional Development Fund under a GameINN project POIR.01.02.00-00-0150/16, operated by The National Centre for Research and Development (NCBiR), and by the Silver Bullet Solutions company. 

\bibliographystyle{IEEEtran}\balance
\bibliography{aaiaDMC17}
\end{document}